\documentclass[anonymous]{article}
\usepackage{ijcai24}

\usepackage{times}
\usepackage{soul}
\usepackage{url}
\usepackage[hidelinks]{hyperref}
\usepackage[utf8]{inputenc}
\usepackage[small]{caption}
\usepackage{graphicx}
\usepackage{amsmath}
\usepackage{amsthm}
\usepackage{booktabs}
\usepackage{algorithm}
\usepackage{algorithmic}
\usepackage[switch]{lineno}
\usepackage{algorithm}
\usepackage{algorithmic}
\usepackage{xcolor}
\usepackage{subcaption}
\usepackage{diagbox}
\usepackage{cancel}
\usepackage{newfloat}
\usepackage{listings}

\title{Compositional Learning of Visually-Grounded Concepts Using Reinforcement}

\author{
Zijun Lin$^1$
\and
Haidi Azaman$^2$\and
M Ganesh Kumar$^3$\and
Cheston Tan$^3$
\affiliations
$^1$Nanyang Technological University\\
$^2$National University of Singapore\\
$^3$Centre for Frontier AI Research, A*STAR
\emails
linz0048@e.ntu.edu.sg,
e0540498@u.nus.edu,
m\_ganeshkumar@u.nus.edu,
cheston\_tan@cfar.a-star.edu.sg
}

\begin{document}

\maketitle

\begin{abstract}

Children can rapidly generalize compositionally-constructed rules to unseen test sets. On the other hand, deep reinforcement learning (RL) agents need to be trained over millions of episodes, and their ability to generalize to unseen combinations remains unclear. Hence, we investigate the compositional abilities of RL agents, using the task of navigating to specified color-shape targets in synthetic 3D environments. First, we show that when RL agents are naively trained to navigate to target color-shape combinations, they implicitly learn to decompose the combinations, allowing them to (re-)compose these and succeed at held-out test combinations (``compositional learning''). Second, when agents are pretrained to learn invariant shape and color concepts (``concept learning''), the number of episodes subsequently needed for compositional learning decreased by 20$\times$. Furthermore, only agents trained on both concept and compositional learning could solve a more complex, out-of-distribution environment in zero-shot fashion. Finally, we verified that only text encoders pretrained on image-text datasets (e.g. CLIP) reduced the number of training episodes needed for our agents to demonstrate compositional learning, and also generalized to 5 unseen colors in zero-shot fashion. Overall, our results are the first to demonstrate that RL agents can be trained to implicitly learn concepts and compositionality, to solve more complex environments in zero-shot fashion. 
 


\end{abstract}

\section{Introduction}

Compositionality is the ability to follow specific rules in putting together basic units of information or primitives \cite{szabo2004compositionality}. To use an informal everyday example, following the instructions from a recipe book, one can prepare a large combination of new food dishes even with a small set of grocery items. Learning to compose basic primitives allows one to generate almost infinitely many solutions to solve complex tasks \cite{todorov2003unsupervised}. However, learning multimodal primitives and composing them to solve a larger number of complex tasks in the vision-language-action domain is a significant gap in existing reinforcement learning (RL) agents \cite{roy2021machine}.

Hierarchical RL agents can learn action primitives and compose them to solve complex tasks~\cite{barto2003recent}, if the agents are sufficiently aided using symbolic methods to explore the sensorimotor space. However, naively training the parameters in RL agents end-to-end requires millions of training episodes \cite{arulkumaran2017deep} and they lack the ability to generalize to out-of-distribution tasks \cite{kirk2023survey}, making them undeployable in the real world \cite{gervet2023navigating}. 


For humans, children first learn individual concepts or schemas \cite{macario1991young,gelman1986categories,piaget1976piaget,rumelhart1977representation} by associating visual and verbal cues, and interacting with their environment \cite{bisk2020experience,iverson2010developing}. \textbf{Learning of concepts in the multi-modal space facilitates subsequent rapid learning of compositional tasks} \cite{mcclelland2013incorporating,kumar2023oneshot}.

Hence, we developed several synthetic 3D environments to train vision-language based navigation agents to associate visual primitives with language-based instructions, by navigating to the correctly referenced object for a reward. Our contributions are as follows:

 \begin{itemize}
     \item We demonstrated the ability of RL agents to learn to decompose color-shape instructions and then (re-)compose them, generalizing to held-out color-shape combinations.

    \item We found that learning of invariant concepts enabled rapid compositional learning of concept combinations, with 100$\times$ and 20$\times$ speed-up for train and test combinations respectively.

    \item We showed for the first time that invariant concept learning that is followed by compositional learning enables the zero-shot ability to handle complex, out-of-distribution compositional tasks.

    \item We found that RL agents with certain pretrained language encoders reduced the number of training episodes needed for compositional learning, and also generalize to out-of-distribution unseen combinations in zero-shot.

 \end{itemize}

\section{Related Work}


\subsection{Environment-grounded multi-modal learning}
Child pedagogies such as Montessori and Reggio emphasize on scaffolding children development using the environment. Specifically, the environment should allow free exploration and contain sufficient resources for children to engage in various stages of play, e.g. solitary, parallel, cooperative, etc.~\cite{parten1932social} while grounding learning experiences to visual and language features.

In recent years, there has been an increase in virtual learning environments to emulate the scaffolding conditions for artificial cognitive development. The simplest form of environment is a two-dimensional continuous arena with boundaries where the agent has to learn state-specific navigation policies to successfully reach targets. The agent perceives its location using place cells and the target to navigate to is given by simple instructions such as a one-hot vector encoding sensory cue \cite{kumar2023oneshot,kumar2022nonlinear}. Due to the simplified task and model requirements, it is easy to understand the learned policies, though hard to transfer to natural conditions. 

A more complex but naturalistic environment is a three-dimensional arena where the agent perceives environmental features using RGB visual inputs, and instructions are represented by either one-hot vectors \cite{Hill} or English sentences \cite{hill2021grounded}. The goal is to navigate to the instructed target. Recent work called XL, allowed the generation of millions of learning environments by changing three variables - the physical 3D space, the game rule specified using natural language, and the number of co-players \cite{team2021open}.

Training RL agents on vastly diverse tasks led to impressive generalization behaviour on held-out tasks, requiring them to navigate, use tools and cooperate or compete depending on the instruction to maximize total rewards. Alternatively, agents can learn to move objects to satisfy the language-based instruction \cite{jiang2019language}. Here, the action space is much larger than a navigation task, as the agent has to learn to manipulate objects to satisfy the corresponding language query. Hence, the visual features are simplified to ensure the learning speed is tractable. 


\subsection{Models for compositional learning}
Compositional learning is the ability to decompose information into its basic elements or concepts, and subsequently combine these concepts to solve novel, unseen combinations in zero-shot or with few examples \cite{lake2015human,xu2021zero,de2006emergence,zuberbuhler2018combinatorial,xia2021temporal,ji2022abstract}. 

Compositional learning has mostly been explored in the context of object detection, where vision-based models are trained on object-attribute pairs, and compose the learned invariances to an unseen test set \cite{kato2018compositional,purushwalkam2019task,anwaar2021compositional}. Alternatively, the loss function can be augmented to encourage networks to decompose information into generalizable features \cite{stone2017teaching,tolooshams2020convolutional}. More recently, models can identify or parse objects from images using bounding boxes~\cite{lee2023determinet} or segmentation masks \cite{kirillov2023segment} to solve new tasks. 

Parsing sentences into individual words and tagging parts-of-speech help to decompose sentences and clarify semantics. Deep networks such as BERT represent concepts with similar or opposite semantics through its embedding~\cite{devlin2018bert,jawahar2019does}, while models trained on large datasets demonstrate compositional reasoning with good performance~\cite{rytting2021leveraging}.

There have been numerous works on learning atomic actions and composing them to rapidly solve new tasks \cite{barto2003recent}. Some of such works are termed hierarchical reinforcement learning where models learn various policies at different levels of control for efficient policy composition \cite{kulkarni2016hierarchical} such as sequential navigation to goals \cite{han2020self}. 

Recent multi-modal models learn to align visual inputs to language concepts \cite{clip,ma2023crepe,yuksekgonul2022and} to solve a multitude of compositional reasoning tasks \cite{lu2023chameleon} such as Visual-Question Answering (VQA) \cite{johnson2017clevr}, Referring Expressions \cite{lee2023determinet}, or augment images using instructions \cite{gupta2023visual}.

Yet, there are only a handful of models that align compositional learning in the vision, language and action spaces. A notable example includes training vision-language based reinforcement learning agents on millions of various toy environments to develop a generally capable agent that can solve tasks based on the instruction given \cite{team2021open}. 

Nevertheless, how these models ground vision-language-action representations for compositional learning, what the individual concepts are, and how these concepts are recomposed to solve novel combinations remains elusive.

\section{Environments for Grounded Learning}

The following section describes the learning environment used to ground the vision-language-navigation reinforcement learning agent on geometric Shapes (S) and Colors (C).

\subsection{Environmental design}

The three 3D environments were developed to learn two key concepts: Shape (S) and Color (C). The environments contain objects made up of five distinct shapes, which are capsule, cube, cylinder, prism, and sphere, and five different colors: red, green, blue, yellow, and black. Hence, each object can be described using the shape and color attributes, for example ``red sphere'', ``blue capsule'', or ``yellow prism''. The Unity engine utilizes the different shape and color combinations described in Table \ref{tab:train test split} to generate new combinations for each episode. 

\begin{figure}
  \centering
  \includegraphics[scale=0.4]{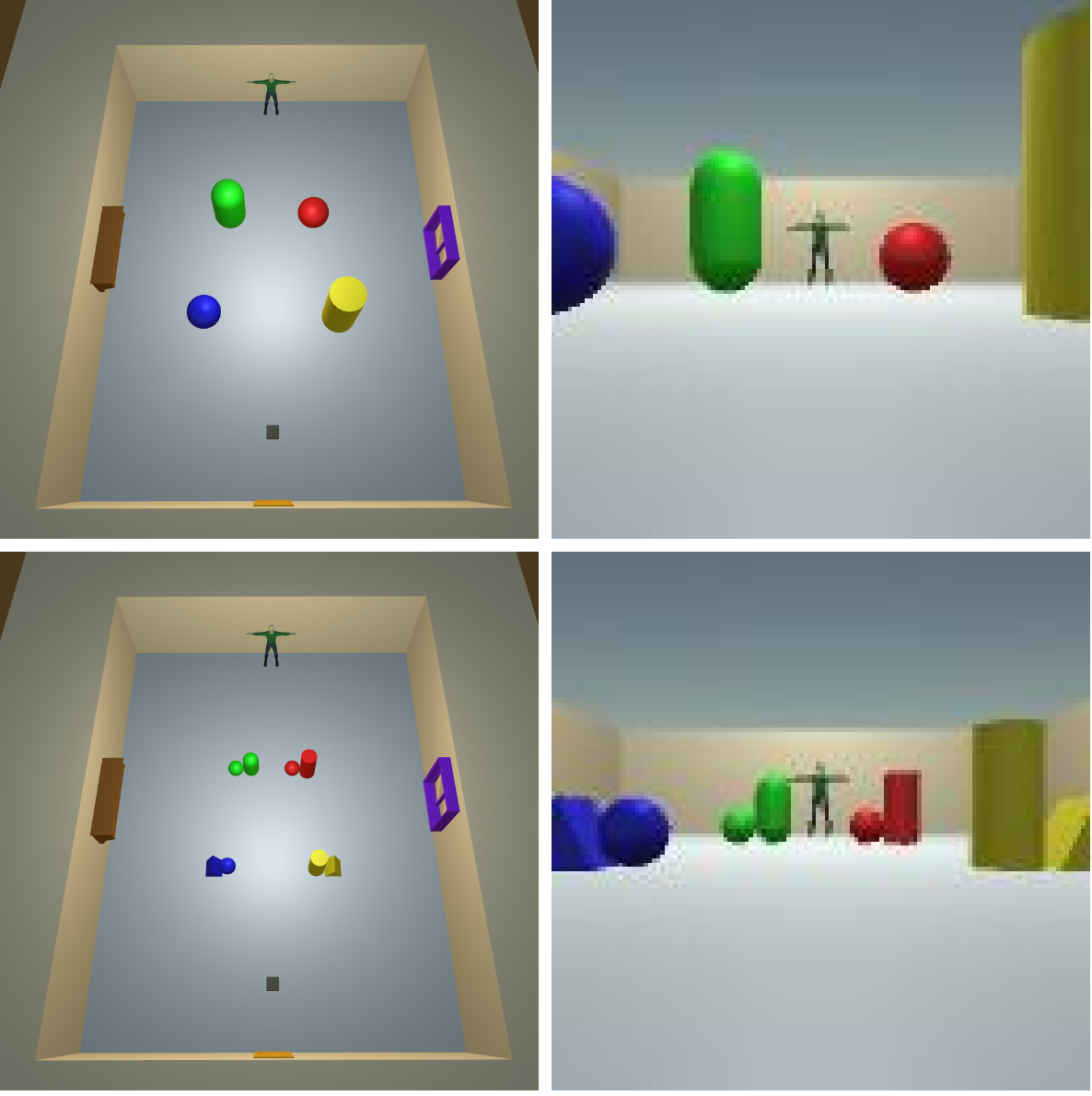}
  \caption{Two example environments. Left column shows the top-view of the environments. Right column shows the RL agent's first-person view (128x128). Top and bottom rows show the C\&S environment (target instruction is ``red sphere'') and the C\&S\&S environment (target instruction is ``red sphere cylinder'') respectively. Notice how the sizes and locations of the objects differ (compare top and bottom).}
  \label{fig:1}
\end{figure}

A target object will be randomly spawned at one of four predetermined locations within a rectangular room. The overall layout of the environment remains constant, as depicted in Figure \ref{fig:1}. The room includes fixed visual cues namely a door, window, shelf and human figure. 

A Unity-based camera models the agent's first-person point-of-view by capturing the environmental dynamics in terms of RGB images. These serve as the visual input for the RL agent. The environment also generates the textual instructions of the target object such as ``red cube'' or ``blue sphere'', which are passed to the RL agent as language input. 

\begin{figure*}
  \centering
  \includegraphics[width=0.8\textwidth,height=0.35\textwidth]{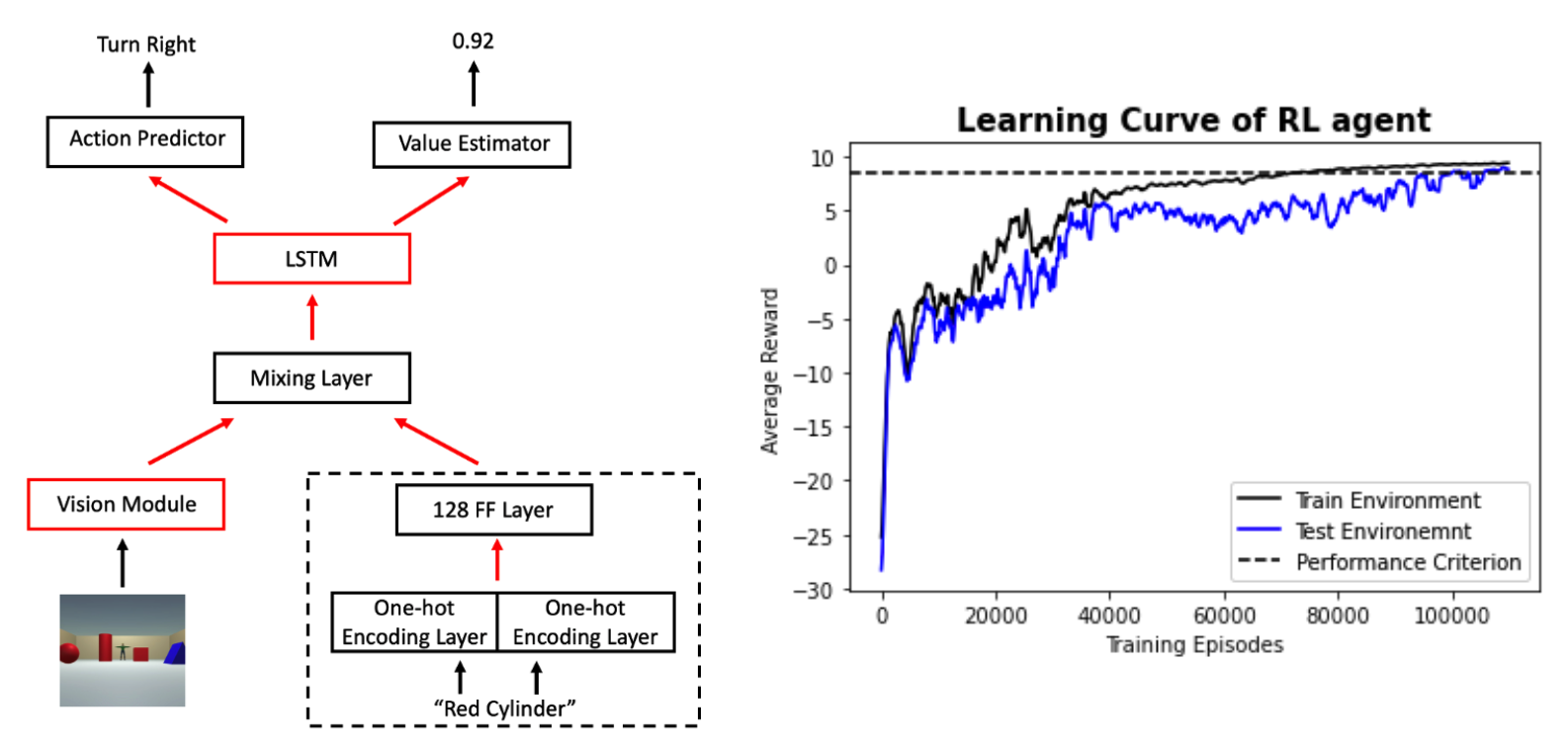}
  \caption{Left: Agent architecture. The language module of the one-hot encoder agent is at bottom right, boxed up within dashed lines. \textbf{\textcolor{red}{Red}} arrows or boxes represent \textbf{\textcolor{red}{trainable}} weights, while \textbf{\textcolor{black}{black}} arrows or boxes represent \textbf{\textcolor{black}{frozen}} weights. Right: Learning curve of the RL agent in train environment and test environments respectively.}\label{fig:agent}
\end{figure*}

\subsection{Task}
\label{sec:Task}
The primary objective of our RL agent is to navigate to the visual object described by the language-based instruction. We devised three different environments to investigate the agent's ability to decompose instructions and learn foundational concepts. The key details of the three environments are: 
 \begin{enumerate}
     \item  C\&S (``Color AND Shape''): The task is to compositionally learn two word concepts. There is one target described by the color and shape attribute with three non-target objects. The instruction includes both color and shape words as inputs.
     
     Additionally, the C*\&S environment (used in Section \ref{sec:exp3}) adds five more color attributes that the agent will not be trained on. This out-of-distribution environment examines the agents' zero-shot capabilities when tested on unseen concepts.
     
     \item  C\&S\&S (``Color AND Shape AND Shape''): The task is to compositionally learn three-word concepts. There is one target with two shapes and three non-targets with pairs of shapes. Both objects in a pair will be of the same color. The instruction includes all three color and two shape words as inputs. This environment evaluates the agents' ability to generalize to out-of-distribution combinations. 
     
      \item  C$|$S (``Color'' OR ``Shape''): The task is to learn single attribute concepts. There is one target and three non-target objects. Only one word, either the color or the shape, is given as instruction. For instance, if the target is ``Blue'', the agent must ignore objects that are non-blue and navigate to the blue object, despite its shape. This environment trains the agent to be invariant to the shape or color respectively. 
 \end{enumerate}

Figure~\ref{fig:1} shows examples of the C\&S (top row) and C\&S\&S (bottom row) environments. For all three environments, one target and three non-target objects or object pairs are randomly spawned at the four pre-defined positions.
\begin{table}
\centering
\begin{tabular}{|c|c|c|c|c|c|}
\hline
\textbf{Shape\textbackslash{}Color} & \textbf{Red} & \textbf{Green} & \textbf{Blue} &\textbf{Yellow} & \textbf{Black} \\
\hline
\textbf{Capsule} & \textbf{\textbf{\textcolor{blue}{Test}}} & Train & Train & Train & Train \\
\hline
\textbf{Cube} & Train & \textbf{\textbf{\textcolor{blue}{Test}}} & Train & Train & Train \\
\hline
\textbf{Cylinder} & Train & Train & \textbf{\textbf{\textcolor{blue}{Test}}} & Train & Train \\
\hline
\textbf{Prism} & Train & Train & Train & \textbf{\textbf{\textcolor{blue}{Test}}} & Train \\
\hline
\textbf{Sphere} & Train & Train & Train & Train & \textbf{\textbf{\textcolor{blue}{Test}}} \\
\hline
\end{tabular}
\caption{Train-Test split for environments C$|$S and C\&S.}
\label{tab:train test split}
\end{table}

Table \ref{tab:train test split} shows the train-test splits for environments C$|$S and C\&S. 20 color-shape combinations were used to train the agents on the relevant tasks and five held-out test combinations were used to evaluate the agents' zero-shot ability to use the rules learned for unseen combinations. The train-test split table for environment C\&S\&S is shown in Supplementary Material.


\subsection{Evaluation Metrics}

During each episode, the environment disburses specific rewards and penalties based on the actions taken by the agent given the state it is in. Successfully navigating to the target object yields a reward of +10, while collisions with non-target objects or walls incur penalties of -3 and -1, respectively. Additionally, the agent receives a penalty of -10 upon reaching the maximum allowed steps of 500. To ascertain the successful learning of a task by the agent, we establish a Performance Criterion of $+$9. The agent is deemed to have effectively learned the task when it accumulates a reward $\geq$ 9 over 100 training episodes.

\section{Agent Architecture for Grounded Learning}

Figure \ref{fig:agent} shows the vision-language agent architecture adapted from \cite{Hill}, which takes in visual inputs (a 3$\times$128$\times$128 tensor of RGB pixel values) and language input (one-hot vector embedding). RGB pixel values are passed into the vision module, which contains three convolutional layers, and the output flattened into a 3136 (a 64$\times$7$\times$7 tensor) dimensional embedding. The language module takes in two one-hot vector embedding of the instructions, each representing the Color and Shape. For example, a complex instruction in C\&S\&S such as ``red cube sphere'' is represented by a one-hot vector ([1,0,0,0,0]) and a two-hot vector ([0,1,0,0,1]). The two vectors are fully connected to a 128 unit linear embedding layer. The 3136-D vector from the vision module and the 128-D vector from the language module are concatenated and fed into a 256-D linear mixing layer. 

A Long Short Term Memory (LSTM) module takes the 256-dimensional embeddings as input. Its activity $s_t$ is passed to both the action predictor (actor) and value estimator (critic). The action predictor maps the LSTM's activity to a probability distribution $\pi(a_t|s_t)$ over four possible actions, i.e., move forward, move backward, turn left and turn right. Meanwhile, the value estimator computes a scalar approximation of the agent's state-value function $V(s_t)$.

The agent is trained using the advantage actor-critic (A2C) algorithm \cite{mnih2016asynchronous,kumar2021one}. We utilize the RMSProp optimizer with a consistent learning rate of $2.5\times10^{-4}$ across all experiments to optimize the training process.

\section{Experiments and Results}

Here, we provide an overview of the experiments and results, which are then detailed in the sub-sections. In \textbf{Experiment 1}, the reinforcement learning agent with a one-hot encoding of instructions was trained on 20 combinations (Table \ref{tab:train test split}) in the C\&S environment to navigate to the color-shape target object. Concurrently, it was evaluated on its zero-shot capability to decompose the five held-out test instructions and navigate to the unseen color-shape target object.

In \textbf{Experiment 2}, we first pretrained agents on the C$|$S environment to learn the concepts of shape and color separately. Only the color or the shape of a target object was given and the agent has to navigate to the corresponding target that is invariant to the second attribute. For example, when given the instruction ``Cube'', the agent has to learn to be invariant to the colors. Surprisingly, pretraining to learn shape or color invariances significantly reduced the number of episodes needed for color-shape compositional learning. We further compare compositional learning with and without concept learning for zero-shot navigation in the out-of-distribution color-shape1-shape2 environment. 

In \textbf{Experiment 3}, we evaluate color-shape compositional learning capabilities of agents with four text-encoder variants, comparing the number of episodes needed to reach the performance criterion. Additionally, we demonstrate their zero-shot capability when presented with out-of-distribution instructions with novel color words.

\subsection{Experiment 1: Generalization of Compositional Learning}

Visually grounded agents can understand single feature instructions \cite{Hill}. How navigation agents learn and compose multiple attributes is unclear. Hence, experiment 1 expands on single attribute navigation to the combination of two attributes, Color $+$ Shape (C\&S). 

To understand whether an RL agent is able to learn to decompose instructions given during training to learn each word group and recompose them to solve color-shape test combinations, the agent is trained on 20 C\&S instructions and tested on 5 held-out C\&S pairs in this experiment as described in Table \ref{tab:train test split}. For example, the agent is trained on the instructions and visual targets ``black cube'' and ``red sphere''. After every 100 training episodes, it is tested on its ability to compose the concepts of ``black'' and ``sphere'' to accurately navigate to the visual target ``black sphere'' when given the held-out test combination instruction. 


In this experiment, three agents were trained ($N=3$) and their mean of the average reward are plotted in Figure~\ref{fig:agent}. The average reward is calculated by taking the average across the most recent 100 episodes after each episode. The learning curve in Figure \ref{fig:agent} shows that the agent with the one-hot text encoder requires approximately 67,000 and 95,000 episodes to achieve performance criterion ($\geq$ 9) for the 20 training and 5 held-out test combinations. This result demonstrates that agents can gradually learn to decompose the color-shape instructions and ground them to the visual attributes during training such that they can recompose the individual concepts to solve the held-out test combinations.

\subsection{Experiment 2: Concept Learning Speeds Up and Generalizes Compositional Learning}

Since the agents are naively trained on the compositional learning task, it is unclear whether the high number of training episodes needed are due to color and shape feature learning, compositional learning or policy learning.

Our hypothesis is that prior learning of the individual Color and Shape attributes \textbf{(Concept learning)} can serve as a foundational schema to enhance the agents' proficiency in learning compositional rules \textbf{(Compositional learning)}. 
To disentangle compositional learning from feature and policy learning, we pretrained the agents to learn the concepts of Color and Shape and learning policies to navigate to the target. The environment of concept learning is C$|$S, in which agents needed 168K episodes to reach the performance criterion (Table \ref{tab:exp2} column 2). Then, these agents were trained on the C\&S environment for compositional learning.

\subsubsection{Concept learning speeds up compositional learning}

Surprisingly, the results in Table \ref{tab:init s2 vs s1 s2} show that the pretrained agents achieved train and held-out test performance criterion in the compositional learning environment 100$\times$ and 20$\times$ faster than naively-trained agents. This suggests that feature and policy learning requires more training episodes than compositional learning. Importantly, pretraining was not only to learn features or policy but to learn the concept of Shape and Color as well.

Concept learning turns out to be more difficult than naively learning to compose as agents need to learn shape and color invariance. For instance, when given the instruction ``black'', the agent needs to navigate to a black object, learning to ignore its shape. Each color or shape has five corresponding shape or color invariants respectively. However, with the instruction ``black cube'', the visual attribute is fixed as both color and shape are specified.



\begin{table}[ht]
    \centering
    \begin{tabular}{|c|c|c|}
        \hline
         & \multicolumn{2}{c|}{\textbf{Episodes (K) for performance criterion}} \\
         \hline
        \multicolumn{1}{|c|}{\textbf{Training}} & \multicolumn{1}{c|}{\centering\textbf{Train}} & \multicolumn{1}{c|}{\centering\textbf{Held-out Test}} \\
        \multicolumn{1}{|c|}{\textbf{Environment}} & \multicolumn{1}{c|}{\centering\textbf{combinations}} & \multicolumn{1}{c|}{\centering\textbf{combinations}} \\
        
        \hline\hspace{0.95cm}
        \textbf{C\&S} & $67.4 \pm 7.2$  & $94.8 \pm 3.7$ \\
        \hline
        \textbf{C$|$S $\rightarrow$ C\&S} & \hspace{0.2cm}$0.6 \pm 0.1$  & \hspace{0.2cm}$5.5 \pm 2.9$ \\
        \hline
    \end{tabular}
    \caption{Mean and standard deviation of the number of episodes (in \textbf{thousands}) required to achieve an average episode reward $\geq$ 9 (performance criterion) over 100 episodes over three repeats. These experiments were run in the C\&S environments, and the results compare agents trained from scratch (i.e. row C\&S), versus with pretraining on C$|$S (i.e. row C$|$S $\rightarrow$ C\&S). Lower values indicate faster learning.}
    \label{tab:init s2 vs s1 s2}
\end{table}

\begin{table*}
    \centering
    \begin{tabular}{|c|r|r|r|r|r|}
        \hline
         & &\multicolumn{4}{c|}{\textbf{Average reward for zero-shot Evaluation in Environments}} \\
        \hline
        \multicolumn{1}{|c|}{\textbf{Training}} & \multicolumn{1}{c|}{\centering\textbf{Training}}& \multicolumn{1}{c|}{\centering\textbf{Familiar}} & \multicolumn{1}{c|}{\centering\textbf{Unseen}}& \multicolumn{1}{c|}{\centering\textbf{Familiar}} & \multicolumn{1}{c|}{\centering\textbf{Unseen}} \\
        \multicolumn{1}{|c|}{\textbf{Environment}} & \multicolumn{1}{c|}{\centering\textbf{Episodes (K)}}& \multicolumn{1}{c|}{\centering\textbf{C$|$S combo}} & \multicolumn{1}{c|}{\centering\textbf{C$|$S combo}} & \multicolumn{1}{c|}{\centering\textbf{C\&S\&S combo}} & \multicolumn{1}{c|}{\centering\textbf{C\&S\&S combo}}\\
        \hline
        \textbf{Nil} & $0.0$&  $-24.42 \pm 1.29$  & $-23.42 \pm 2.57$ & $-29.15 \pm 3.07$ & $-36.48 \pm 3.60$\\
        \hline
        \multicolumn{1}{|r|}{\textbf{C\&S}}& $67.4$  & $0.37 \pm 1.50$ & $3.08 \pm 0.32$ & $2.84 \pm 0.92$ & $-5.10 \pm 2.59$ \\
        \hline
        \multicolumn{1}{|r|}{\textbf{C\&S}}& $168.6$ & $-8.02 \pm 3.01$ & $-2.05 \pm 1.57$ & $-8.27 \pm 3.75$ & $-23.2 \pm 9.97$ \\
        \hline
         \multicolumn{0}{|l|}{\hspace{0.02cm}\textbf{C$|$S}}& {$168\hspace{1cm}$}& \diagbox[width=2.6cm, height=0.4cm]{}{}  & \diagbox[width=2.6cm, height=0.4cm]{}{} & $1.19 \pm 1.24$ & $-4.02 \pm 2.44$\\
        \hline
        \textbf{C$|$S\hspace{0.06cm} $\rightarrow$ \hspace{0.06cm}C\&S}& $168 \rightarrow 0.6$ & $\mathbf{8.74 \pm 0.29}$ & $\mathbf{7.58 \pm 0.32}$& $\mathbf{5.49 \pm 0.26}$ & $\mathbf{5.55\pm 0.39}$\\
        \hline
    \end{tabular}
    \caption{Summary of zero-shot evaluation experiments. The values are the mean and standard deviation of the rewards obtained by ($N=3$) agents over 100 episodes in the novel environments with color and shape combinations previously trained on (familiar) and untrained on (unseen). Higher reward values indicate better performance.}
    \label{tab:exp2}
\end{table*}

\subsubsection{Concept and compositional learning improves generalization}
So far, agents have only been trained and tested on color and shape combinations. However, in the real-world, instructions are highly compositional, going beyond two word color-shape phrases. Furthermore, real-world objects can be composed of two or more basic shapes, such as a hammer being composed of a cylinder and a cuboid. 

To investigate the generalizability of the agents, we created the C\&S\&S test-only environment (Figure~\ref{fig:1}, bottom) where RL agents have to compose three word instructions, one color and two shapes and navigate to the correct target which is composed of two objects, making the task more complex and difficult than C\&S. Importantly, the RL agents are not trained on the C\&S\&S evaluation environment. Since there are some overlap between the train combinations in C\&S (Table \ref{tab:train test split}) and the evaluation combinations in C\&S\&S (Supplementary Table 5), these objects and instructions are grouped into Familiar Combinations (Table \ref{tab:exp2}, column 5), while C\&S\&S evaluation combinations that do not overlap with the C\&S train combinations are called Unseen Combinations (Table \ref{tab:exp2}, column 6), making them truly novel and out-of-distribution composed objects and instructions. 


Table \ref{tab:exp2} shows the zero-shot evaluation performance on the C\&S\&S environment of agents trained either on concept learning (C$|$S), compositional learning (C\&S) or both (C$|$S $\rightarrow$ C\&S). For the agent exclusively trained on compositional learning, two training checkpoints were selected based on training episodes. The first is the 67.4K checkpoint at which the agent achieved performance criterion. The second checkpoint is at 168.6K which is an agent that is trained for the same number of training episodes as agents undergoing both concept and subsequently compositional learning (last row).



The best performing agents were the ones that were trained on C$|$S first and then C\&S, achieving rewards of 5.5 on both the familiar and unseen C\&S\&S combinations in zero-shot. None of the other agents demonstrated equivalent zero-shot proficiency in the C\&S\&S environment, even if the agent was trained for the same number of training episodes on the compositional learning task. Interestingly, even though the agent was pretrained on C$|$S and then trained on C\&S, it maintained its zero-shot performance on the C$|$S combinations, implying that its foundational conceptual knowledge of colors and shape was not forgotten. These results suggest that out-of-distribution generalization cannot be obtained by simply learning features, composition and navigation policy, but rather foundational concepts need to be learned first before learning to compose them to solve more complex tasks.  

Furthermore, Table \ref{tab:exp2} shows that compared to the agent trained on C\&S for 67.4K episodes, agents trained only on C$|$S perform slightly worse on the familiar C\&S\&S combinations, but better on the unseen C\&S\&S combinations. While these agents' zero-shot generalization performance are nowhere close to the sequentially-trained agent's performance (C$|$S$\rightarrow$C\&S), they perform much better than the agents that were over-trained in C\&S for 168.6K episodes, as well as the agents that were not trained in any environment so as to show chance performance (``Nil'').




\subsubsection{Representation learning through concept learning}
\begin{figure}[H]
  \centering
   \includegraphics[width=0.92\columnwidth]{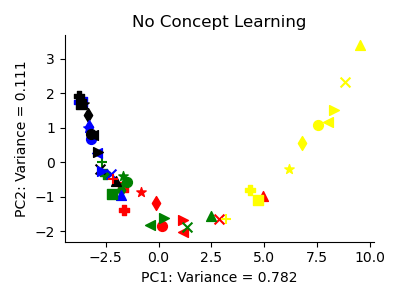}
  \hspace{0.001in}
   \includegraphics[width=0.92\columnwidth]{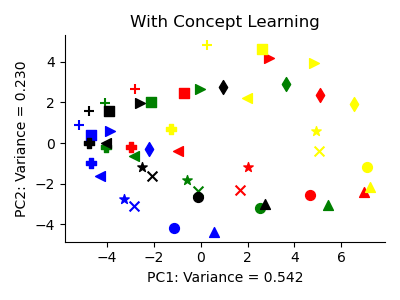}
  \caption{Embedding of average LSTM activity for 50 instructions from  C\&S\&S after training only on C\&S (Top, ``No Concept Learning''), compared to pretraining on C$|$S and then on C\&S (Bottom, ``With Concept Learning'').}
  \label{app:conceptlearn}
\end{figure}

Figure \ref{app:conceptlearn} demonstrates the differences in embedding the 50 combinations in the C\&S\&S environments by the LSTM layer. When an agent does not explicitly learn the color and shape concepts and instead only learns to compose, the agent learns a low dimensional representation (three Principal Components) to represent 90\% of its information. Hence, the agent may not have sufficient expressivity to represent and manipulate more complex combinations in its low dimensional space. 

Conversely, learning concepts and then learning to compose allows the LSTM to represent the C\&S\&S combinations in a slightly larger representation space (seven Principal Components) to encode 90\% of the information. The higher expressivity offered by a larger representation space could be a reason why the agent is able to map more complex combinations for manipulation. 




\subsection{Experiment 3: Comparison of Text Encoders}\label{sec:exp3}

Thus far, we have seen an agent with a vanilla visual module and one-hot encoded text module can be trained end-to-end using a reinforcement learning objective. We next hypothesized that substituting vanilla one-hot text module with pretrained but frozen text encoders such as CLIP \cite{clip} and BERT \cite{bert} might reduce the number of training episodes needed to achieve performance criterion. Additionally, we constructed a vanilla text encoders to serve as a baseline. This encoder utilizes the CLIP tokenizer, and the tokens are passed through an embedding layer, a pooling layer and a feed-forward layer. The parameters of the vanilla text encoder are trained end-to-end like the one-hot encoder.

Details of agent architecture with different pretrained text encoders are in the Supplementary Material. The only changes made to the RL agent is to substitute the one-hot encoder with pretrained or vanilla text encoders. Table \ref{tab:text encoder results} shows the average number of training episodes needed for three agents ($N=3$) initialized with different text encoders to achieve performance criterion in the C\&S environment on both the train and test combinations.

\begin{table}[ht]
    \centering
    \begin{tabular}{|c|r|r|}
        \hline
         & \multicolumn{2}{c|}{\textbf{Training Episodes (K)}} \\
        \hline
            \multicolumn{1}{|c|}{\textbf{Text}} & \multicolumn{1}{c|}{\centering\textbf{Train}} & \multicolumn{1}{c|}{\centering\textbf{Test}} \\
            \multicolumn{1}{|c|}{\textbf{Encoder}} & \multicolumn{1}{c|}{\centering\textbf{combinations}} & \multicolumn{1}{c|}{\centering\textbf{combinations}} \\
        \hline
        \textbf{One-hot} & $67.4 \pm 7.2$  & $94.8 \pm 3.7$ \\
        \hline
         \textbf{Vanilla} & $116.2 \pm 15.4$ & $185.9 \pm 15.5$ \\
        \hline
        \textbf{BERT} & $109.0 \pm 9.1$ & $\geq 200$\\
        \hline
        \textbf{CLIP} & $\mathbf{56.2 \pm 5.3}$ & $\mathbf{72.6 \pm 6.0}$ \\
        \hline
    \end{tabular}
    \caption{Init $\rightarrow$ C\&S: Learning to decompose instructions for compositional performance. Values in the table indicate the mean of episodes (in \textbf{thousands}) needed to achieve an average episode reward $\geq$ 9 (performance criterion) over 100 episodes with standard deviation across three repeats in training and testing environments. Lower values indicate faster learning.}
    \label{tab:text encoder results}
\end{table}

The vanilla text encoder used CLIP's tokenizer to encode the various colors and shapes into orthogonal tokens, similar to the one-hot encoder, making it easier for the vanilla text encoder to disentangle the embedding even before training (refer to Supplementary material). However, training from scratch required about two times the training episodes of the one-hot encoder to achieve performance criterion for the train and test combinations. The longer training episodes could be due to the larger number of parameters in the embedding layer, which is absent in the one-hot text encoder. 

\begin{figure}[H]
  \centering
   \includegraphics[width=\columnwidth]{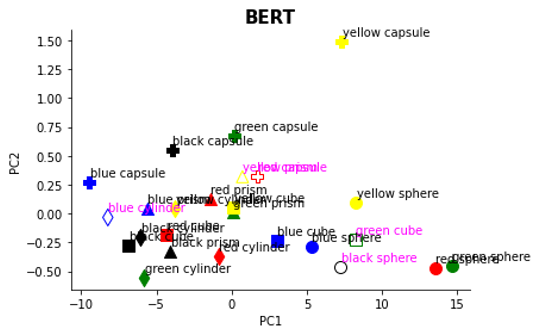}
  \hspace{0.001in}
   \includegraphics[width=\columnwidth]{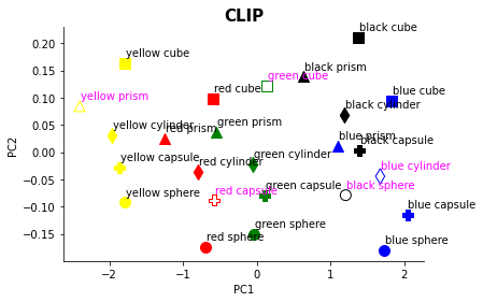}
  \caption{Word embeddings of the agent with BERT (top) and CLIP (bottom) text encoders after training in the C\&S environment for 50K episodes. Filled icons represent training set examples, while unfilled icons with magenta labels represent testing set examples.}
  \label{fig:word_embeddings}
\end{figure}

Although the agent with the BERT text encoder was trained for a maximum of 200,000 episodes, the agent only achieved an average reward of 8.5, failing to reach performance criterion for testing combinations. Figure \ref{fig:word_embeddings} shows that after training for 50,000 episodes, BERT's word embedding of the 20 train and 5 test instructions remain highly overlapped, suggesting that the pretrained BERT encoder struggles to distinguish between the color and shape concepts. A potential explanation is because BERT is only trained on textual data and its representations are not grounded to visual attributes, making it difficult to disambiguate and associate to visual concepts like color and shape. 
The CLIP text encoder achieved performance criterion on both the train and test combinations 1.3 times faster than the one-hot text encoder agent. The expedited learning implies that CLIP's prelearned word embedding is useful to increase the training efficiency for a reinforcement learning agent. Figure~\ref{fig:word_embeddings} shows that CLIP encodes both the train and test instructions systematically where Principal Component 1 and 2 clearly explains the color and shape dimensions with minimal overlap between the 25 instructions.

Such organization of color and shape concepts could be because CLIP's language comprehension is grounded to visual attributes due to its cross-modal training regime on various image-caption datasets. This result demonstrates that it is possible to use vision-language grounded models to improve the training efficiency of multi-modal reinforcement learning agents, especially for compositional learning.

\subsubsection{Out-of-distribution generalisation capability}

To assess the agent's ability to generalize to out-of-distribution instructions not encountered during training, the agent with CLIP text encoder was tested in the C*\&S environment (see Section~\ref{sec:Task}), where five new colors (Orange, Cyan, Pink, Purple, and White) were introduced. Example instructions were ``cyan prism'' or ``purple cube'', but ``cyan'' and ``purple'' were never presented during training.

\textbf{The agent achieves a noteworthy zero-shot result, averaging 6.9 over three repeats (100 test episodes at each repeat)}, demonstrating that agents trained on the original five colors, equipped with a CLIP text encoder, successfully adapted to the five novel colors in the C*\&S environment. Thus, the abilities of these agents are not limited to the 25 color and shape combinations, and can comprehend words not encountered previously.

\section{Limitations and Future work}

The 3D environments developed for this study utilized relatively basic geometric shapes such as ``capsule'' and ``prism''. This raises questions about generalization to more realistic objects. Also, without any obstacles in the room, the navigation itself is simple. As such, an agent trained in our environments may not generalize to other scenarios with more complex navigation requirement. 

In the concept learning experiment, we only compared the performance of the one-hot agent across different training and testing environments. As such, we did not provide insights on whether pretrained and frozen visual encoders can show similar increases in speed to solve compositional learning tasks. 

Furthermore, the language instructions for our experiments are limited to just two or three words. All of these limitations will be explored as future work.



\section{Conclusion}

We demonstrated the compositional abilities of reinforcement learning agents. Specifically, we found that agents can learn to decompose and recompose instructions to solve held-out test instructions. Furthermore, we showed that invariant concept learning accelerates compositional learning.

We evaluated the zero-shot capabilities of agents on novel and complex compositional learning tasks after learning concepts and compositionality using a smaller training set. Finally, we tested various text encoders, with the CLIP text encoder demonstrating the ability to speed up learning. CLIP even enables the ability to follow unseen, out-of-distribution instructions.



\bibliographystyle{named}
\bibliography{ijcai24.bib}

\begin{thebibliography}{}

\bibitem[\protect\citeauthoryear{Anwaar \bgroup \em et al.\egroup }{2021}]{anwaar2021compositional}
Muhammad~Umer Anwaar, Egor Labintcev, and Martin Kleinsteuber.
\newblock Compositional learning of image-text query for image retrieval.
\newblock In {\em Proceedings of the IEEE/CVF Winter conference on Applications of Computer Vision}, pages 1140--1149, 2021.

\bibitem[\protect\citeauthoryear{Arulkumaran \bgroup \em et al.\egroup }{2017}]{arulkumaran2017deep}
Kai Arulkumaran, Marc~Peter Deisenroth, Miles Brundage, and Anil~Anthony Bharath.
\newblock Deep reinforcement learning: A brief survey.
\newblock {\em IEEE Signal Processing Magazine}, 34(6):26--38, 2017.

\bibitem[\protect\citeauthoryear{Barto and Mahadevan}{2003}]{barto2003recent}
Andrew~G Barto and Sridhar Mahadevan.
\newblock Recent advances in hierarchical reinforcement learning.
\newblock {\em Discrete event dynamic systems}, 13(1-2):41--77, 2003.

\bibitem[\protect\citeauthoryear{Bisk \bgroup \em et al.\egroup }{2020}]{bisk2020experience}
Yonatan Bisk, Ari Holtzman, Jesse Thomason, Jacob Andreas, Yoshua Bengio, Joyce Chai, Mirella Lapata, Angeliki Lazaridou, Jonathan May, Aleksandr Nisnevich, et~al.
\newblock Experience grounds language.
\newblock {\em arXiv preprint arXiv:2004.10151}, 2020.

\bibitem[\protect\citeauthoryear{De~Beule and Bergen}{2006}]{de2006emergence}
Joachim De~Beule and Benjamin~K Bergen.
\newblock On the emergence of compositionality.
\newblock In {\em The Evolution of Language}, pages 35--42. World Scientific, 2006.

\bibitem[\protect\citeauthoryear{Devlin \bgroup \em et al.\egroup }{2018}]{devlin2018bert}
Jacob Devlin, Ming-Wei Chang, Kenton Lee, and Kristina Toutanova.
\newblock Bert: Pre-training of deep bidirectional transformers for language understanding.
\newblock {\em arXiv preprint arXiv:1810.04805}, 2018.

\bibitem[\protect\citeauthoryear{Devlin \bgroup \em et al.\egroup }{2019}]{bert}
Jacob Devlin, Ming-Wei Chang, Kenton Lee, and Kristina Toutanova.
\newblock {BERT}: Pre-training of deep bidirectional transformers for language understanding.
\newblock In {\em Proceedings of the 2019 Conference of the North {A}merican Chapter of the Association for Computational Linguistics: Human Language Technologies, Volume 1 (Long and Short Papers)}, Minneapolis, Minnesota, June 2019.

\bibitem[\protect\citeauthoryear{Gelman and Markman}{1986}]{gelman1986categories}
Susan~A Gelman and Ellen~M Markman.
\newblock Categories and induction in young children.
\newblock {\em Cognition}, 23(3):183--209, 1986.

\bibitem[\protect\citeauthoryear{Gervet \bgroup \em et al.\egroup }{2023}]{gervet2023navigating}
Theophile Gervet, Soumith Chintala, Dhruv Batra, Jitendra Malik, and Devendra~Singh Chaplot.
\newblock Navigating to objects in the real world.
\newblock {\em Science Robotics}, 8(79):eadf6991, 2023.

\bibitem[\protect\citeauthoryear{Gupta and Kembhavi}{2023}]{gupta2023visual}
Tanmay Gupta and Aniruddha Kembhavi.
\newblock Visual programming: Compositional visual reasoning without training.
\newblock In {\em Proceedings of the IEEE/CVF Conference on Computer Vision and Pattern Recognition}, pages 14953--14962, 2023.

\bibitem[\protect\citeauthoryear{Han \bgroup \em et al.\egroup }{2020}]{han2020self}
Dongqi Han, Kenji Doya, and Jun Tani.
\newblock Self-organization of action hierarchy and compositionality by reinforcement learning with recurrent neural networks.
\newblock {\em Neural Networks}, 129:149--162, 2020.

\bibitem[\protect\citeauthoryear{Hill \bgroup \em et al.\egroup }{2020}]{Hill}
Felix Hill, Stephen Clark, Phil Blunsom, and Karl~Moritz Hermann.
\newblock Simulating early word learning in situated connectionist agents.
\newblock In {\em Annual Meeting of the Cognitive Science Society}, 2020.

\bibitem[\protect\citeauthoryear{Hill \bgroup \em et al.\egroup }{2021}]{hill2021grounded}
Felix Hill, Olivier Tieleman, Tamara Von~Glehn, Nathaniel Wong, Hamza Merzic, and Stephen Clark.
\newblock Grounded language learning fast and slow.
\newblock {\em ICLR}, 2021.

\bibitem[\protect\citeauthoryear{Iverson}{2010}]{iverson2010developing}
Jana~M Iverson.
\newblock Developing language in a developing body: The relationship between motor development and language development.
\newblock {\em Journal of child language}, 37(2):229--261, 2010.

\bibitem[\protect\citeauthoryear{Jawahar \bgroup \em et al.\egroup }{2019}]{jawahar2019does}
Ganesh Jawahar, Beno{\^\i}t Sagot, and Djam{\'e} Seddah.
\newblock What does bert learn about the structure of language?
\newblock In {\em ACL 2019-57th Annual Meeting of the Association for Computational Linguistics}, 2019.

\bibitem[\protect\citeauthoryear{Ji \bgroup \em et al.\egroup }{2022}]{ji2022abstract}
Anya Ji, Noriyuki Kojima, Noah Rush, Alane Suhr, Wai~Keen Vong, Robert~D Hawkins, and Yoav Artzi.
\newblock Abstract visual reasoning with tangram shapes.
\newblock {\em arXiv preprint arXiv:2211.16492}, 2022.

\bibitem[\protect\citeauthoryear{Jiang \bgroup \em et al.\egroup }{2019}]{jiang2019language}
Yiding Jiang, Shixiang~Shane Gu, Kevin~P Murphy, and Chelsea Finn.
\newblock Language as an abstraction for hierarchical deep reinforcement learning.
\newblock {\em Advances in Neural Information Processing Systems}, 32, 2019.

\bibitem[\protect\citeauthoryear{Johnson \bgroup \em et al.\egroup }{2017}]{johnson2017clevr}
Justin Johnson, Bharath Hariharan, Laurens Van Der~Maaten, Li~Fei-Fei, C~Lawrence~Zitnick, and Ross Girshick.
\newblock Clevr: A diagnostic dataset for compositional language and elementary visual reasoning.
\newblock In {\em Proceedings of the IEEE conference on computer vision and pattern recognition}, pages 2901--2910, 2017.

\bibitem[\protect\citeauthoryear{Kato \bgroup \em et al.\egroup }{2018}]{kato2018compositional}
Keizo Kato, Yin Li, and Abhinav Gupta.
\newblock Compositional learning for human object interaction.
\newblock In {\em Proceedings of the European Conference on Computer Vision (ECCV)}, pages 234--251, 2018.

\bibitem[\protect\citeauthoryear{Kirillov \bgroup \em et al.\egroup }{2023}]{kirillov2023segment}
Alexander Kirillov, Eric Mintun, Nikhila Ravi, Hanzi Mao, Chloe Rolland, Laura Gustafson, Tete Xiao, Spencer Whitehead, Alexander~C Berg, Wan-Yen Lo, et~al.
\newblock Segment anything.
\newblock {\em arXiv preprint arXiv:2304.02643}, 2023.

\bibitem[\protect\citeauthoryear{Kirk \bgroup \em et al.\egroup }{2023}]{kirk2023survey}
Robert Kirk, Amy Zhang, Edward Grefenstette, and Tim Rockt{\"a}schel.
\newblock A survey of zero-shot generalisation in deep reinforcement learning.
\newblock {\em Journal of Artificial Intelligence Research}, 76:201--264, 2023.

\bibitem[\protect\citeauthoryear{Kulkarni \bgroup \em et al.\egroup }{2016}]{kulkarni2016hierarchical}
Tejas~D Kulkarni, Karthik Narasimhan, Ardavan Saeedi, and Josh Tenenbaum.
\newblock Hierarchical deep reinforcement learning: Integrating temporal abstraction and intrinsic motivation.
\newblock {\em Advances in neural information processing systems}, 29, 2016.

\bibitem[\protect\citeauthoryear{Kumar \bgroup \em et al.\egroup }{2021}]{kumar2021one}
M~Ganesh Kumar, Cheston Tan, Camilo Libedinsky, Shih-Cheng Yen, and Andrew Yong-Yi Tan.
\newblock One-shot learning of paired associations by a reservoir computing model with hebbian plasticity.
\newblock {\em arXiv preprint arXiv:2106.03580}, 2021.

\bibitem[\protect\citeauthoryear{Kumar \bgroup \em et al.\egroup }{2022}]{kumar2022nonlinear}
M~Ganesh Kumar, Cheston Tan, Camilo Libedinsky, Shih-Cheng Yen, and Andrew~YY Tan.
\newblock A nonlinear hidden layer enables actor--critic agents to learn multiple paired association navigation.
\newblock {\em Cerebral Cortex}, 32(18):3917--3936, 2022.

\bibitem[\protect\citeauthoryear{Kumar \bgroup \em et al.\egroup }{2023}]{kumar2023oneshot}
M~Ganesh Kumar, Cheston Tan, Camilo Libedinsky, Shih-Cheng Yen, and Andrew Yong-Yi Tan.
\newblock One-shot learning of paired association navigation with biologically plausible schemas.
\newblock {\em arXiv preprint arXiv:2205.09710}, 2023.

\bibitem[\protect\citeauthoryear{Lake \bgroup \em et al.\egroup }{2015}]{lake2015human}
Brenden~M Lake, Ruslan Salakhutdinov, and Joshua~B Tenenbaum.
\newblock Human-level concept learning through probabilistic program induction.
\newblock {\em Science}, 350(6266):1332--1338, 2015.

\bibitem[\protect\citeauthoryear{Lee \bgroup \em et al.\egroup }{2023}]{lee2023determinet}
Clarence Lee, M~Ganesh Kumar, and Cheston Tan.
\newblock Determinet: A large-scale diagnostic dataset for complex visually-grounded referencing using determiners.
\newblock {\em ICCV}, 2023.

\bibitem[\protect\citeauthoryear{Lu \bgroup \em et al.\egroup }{2023}]{lu2023chameleon}
Pan Lu, Baolin Peng, Hao Cheng, Michel Galley, Kai-Wei Chang, Ying~Nian Wu, Song-Chun Zhu, and Jianfeng Gao.
\newblock Chameleon: Plug-and-play compositional reasoning with large language models.
\newblock {\em arXiv preprint arXiv:2304.09842}, 2023.

\bibitem[\protect\citeauthoryear{Ma \bgroup \em et al.\egroup }{2023}]{ma2023crepe}
Zixian Ma, Jerry Hong, Mustafa~Omer Gul, Mona Gandhi, Irena Gao, and Ranjay Krishna.
\newblock Crepe: Can vision-language foundation models reason compositionally?
\newblock In {\em Proceedings of the IEEE/CVF Conference on Computer Vision and Pattern Recognition}, pages 10910--10921, 2023.

\bibitem[\protect\citeauthoryear{Macario}{1991}]{macario1991young}
Jason~F Macario.
\newblock Young children's use of color in classification: Foods and canonically colored objects.
\newblock {\em Cognitive Development}, 6(1):17--46, 1991.

\bibitem[\protect\citeauthoryear{McClelland}{2013}]{mcclelland2013incorporating}
James~L McClelland.
\newblock Incorporating rapid neocortical learning of new schema-consistent information into complementary learning systems theory.
\newblock {\em Journal of Experimental Psychology: General}, 142(4):1190, 2013.

\bibitem[\protect\citeauthoryear{Mnih \bgroup \em et al.\egroup }{2016}]{mnih2016asynchronous}
Volodymyr Mnih, Adria~Puigdomenech Badia, Mehdi Mirza, Alex Graves, Timothy Lillicrap, Tim Harley, David Silver, and Koray Kavukcuoglu.
\newblock Asynchronous methods for deep reinforcement learning.
\newblock In {\em International conference on machine learning}, pages 1928--1937. PMLR, 2016.

\bibitem[\protect\citeauthoryear{Parten}{1932}]{parten1932social}
Mildred~B Parten.
\newblock Social participation among pre-school children.
\newblock {\em The Journal of Abnormal and Social Psychology}, 27(3):243, 1932.

\bibitem[\protect\citeauthoryear{Piaget}{1976}]{piaget1976piaget}
Jean Piaget.
\newblock Piaget’s theory, 1976.

\bibitem[\protect\citeauthoryear{Purushwalkam \bgroup \em et al.\egroup }{2019}]{purushwalkam2019task}
Senthil Purushwalkam, Maximilian Nickel, Abhinav Gupta, and Marc'Aurelio Ranzato.
\newblock Task-driven modular networks for zero-shot compositional learning.
\newblock In {\em Proceedings of the IEEE/CVF International Conference on Computer Vision}, pages 3593--3602, 2019.

\bibitem[\protect\citeauthoryear{Radford \bgroup \em et al.\egroup }{2021}]{clip}
Alec Radford, Jong~Wook Kim, Chris Hallacy, Aditya Ramesh, Gabriel Goh, Sandhini Agarwal, Girish Sastry, Amanda Askell, Pamela Mishkin, Jack Clark, Gretchen Krueger, and Ilya Sutskever.
\newblock Learning transferable visual models from natural language supervision, 2021.

\bibitem[\protect\citeauthoryear{Roy \bgroup \em et al.\egroup }{2021}]{roy2021machine}
Nicholas Roy, Ingmar Posner, Tim Barfoot, Philippe Beaudoin, Yoshua Bengio, Jeannette Bohg, Oliver Brock, Isabelle Depatie, Dieter Fox, Dan Koditschek, et~al.
\newblock From machine learning to robotics: challenges and opportunities for embodied intelligence.
\newblock {\em arXiv preprint arXiv:2110.15245}, 2021.

\bibitem[\protect\citeauthoryear{Rumelhart and Ortony}{1977}]{rumelhart1977representation}
David~E Rumelhart and Andrew Ortony.
\newblock The representation of knowledge in memory.
\newblock {\em Schooling and the acquisition of knowledge}, 99:135, 1977.

\bibitem[\protect\citeauthoryear{Rytting and Wingate}{2021}]{rytting2021leveraging}
Christopher Rytting and David Wingate.
\newblock Leveraging the inductive bias of large language models for abstract textual reasoning.
\newblock {\em Advances in Neural Information Processing Systems}, 34:17111--17122, 2021.

\bibitem[\protect\citeauthoryear{Stone \bgroup \em et al.\egroup }{2017}]{stone2017teaching}
Austin Stone, Huayan Wang, Michael Stark, Yi~Liu, D~Scott~Phoenix, and Dileep George.
\newblock Teaching compositionality to cnns.
\newblock In {\em Proceedings of the IEEE conference on computer vision and pattern recognition}, pages 5058--5067, 2017.

\bibitem[\protect\citeauthoryear{Szab\'o}{2008}]{szabo2004compositionality}
Zolt\'{a}n~Gendler Szab\'o.
\newblock Compositionality.
\newblock In {\em Stanford Encyclopedia of Philosophy}. 2008.

\bibitem[\protect\citeauthoryear{Team \bgroup \em et al.\egroup }{2021}]{team2021open}
Open Ended~Learning Team, Adam Stooke, Anuj Mahajan, Catarina Barros, Charlie Deck, Jakob Bauer, Jakub Sygnowski, Maja Trebacz, Max Jaderberg, Michael Mathieu, et~al.
\newblock Open-ended learning leads to generally capable agents.
\newblock {\em arXiv preprint arXiv:2107.12808}, 2021.

\bibitem[\protect\citeauthoryear{Todorov and Ghahramani}{2003}]{todorov2003unsupervised}
Emanuel Todorov and Zoubin Ghahramani.
\newblock Unsupervised learning of sensory-motor primitives.
\newblock In {\em Proceedings of the 25th Annual International Conference of the IEEE Engineering in Medicine and Biology Society (IEEE Cat. No. 03CH37439)}, volume~2, pages 1750--1753. IEEE, 2003.

\bibitem[\protect\citeauthoryear{Tolooshams \bgroup \em et al.\egroup }{2020}]{tolooshams2020convolutional}
Bahareh Tolooshams, Andrew Song, Simona Temereanca, and Demba Ba.
\newblock Convolutional dictionary learning based auto-encoders for natural exponential-family distributions.
\newblock In {\em International Conference on Machine Learning}, pages 9493--9503. PMLR, 2020.

\bibitem[\protect\citeauthoryear{Xia and Collins}{2021}]{xia2021temporal}
Liyu Xia and Anne~GE Collins.
\newblock Temporal and state abstractions for efficient learning, transfer, and composition in humans.
\newblock {\em Psychological review}, 128(4):643, 2021.

\bibitem[\protect\citeauthoryear{Xu \bgroup \em et al.\egroup }{2021}]{xu2021zero}
Guangyue Xu, Parisa Kordjamshidi, and Joyce~Y Chai.
\newblock Zero-shot compositional concept learning.
\newblock {\em arXiv preprint arXiv:2107.05176}, 2021.

\bibitem[\protect\citeauthoryear{Yuksekgonul \bgroup \em et al.\egroup }{2022}]{yuksekgonul2022and}
Mert Yuksekgonul, Federico Bianchi, Pratyusha Kalluri, Dan Jurafsky, and James Zou.
\newblock When and why vision-language models behave like bags-of-words, and what to do about it?
\newblock In {\em The Eleventh International Conference on Learning Representations}, 2022.

\bibitem[\protect\citeauthoryear{Zuberb{\"u}hler}{2018}]{zuberbuhler2018combinatorial}
Klaus Zuberb{\"u}hler.
\newblock Combinatorial capacities in primates.
\newblock {\em Current Opinion in Behavioral Sciences}, 21:161--169, 2018.

\end{thebibliography}
\newpage
\clearpage

\section*{Supplementary Materials}
\vspace{12pt}

\section{Agent Architectures with various text encoders}

\begin{figure}[ht]
  \centering
   \includegraphics[width=0.45\textwidth,height=0.25\textwidth]{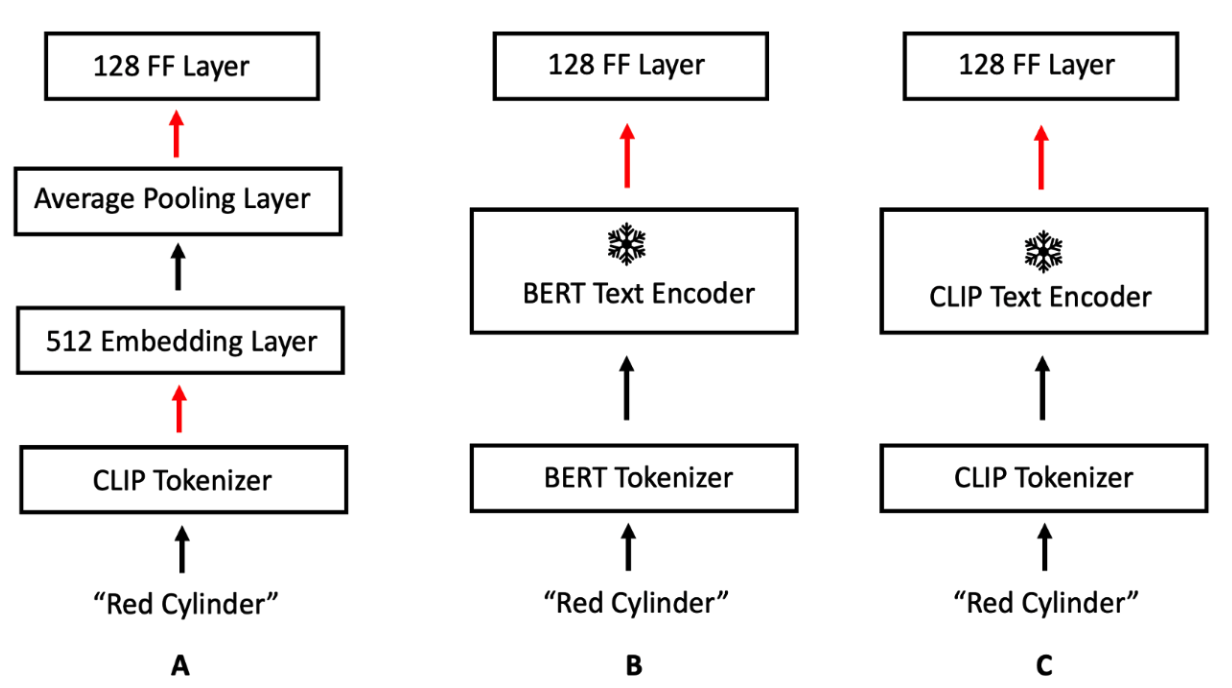}
   \caption{The language model of the agent with:  \textbf{A}) Vanilla text encoder,  \textbf{B}) BERT text encoder, \textbf{C}) CLIP text encoder are shown respectively. BERT text encoder and CLIP text encoder are both pretrained and frozen. Red arrows or boxes represent trainable weights, black arrows or boxes represent frozen weights.}
  \label{fig:te}
\end{figure}

To study the influence of different text encoders on learning word combinations, we introduce three variants: vanilla, CLIP, and BERT, whose architectures are depicted in Figure \ref{fig:te} respectively. The output dimensions of three text encoders are standardized to 128 for fair comparison. Instructions are represented using text strings and passed to the language modules. Importantly, the other parts of the agent is unchanged to be consistent with the One-hot agent.

\subsubsection{Vanilla Text Encoder}

The text instruction firstly undergoes tokenization using the original CLIP tokenizer, with a maximum tokenized tensor length of 77 \cite{clip}. The tokenized tensor is then passed through a single token embedding layer, retrieving 512-dimensional word embeddings for each token. Subsequently, an average pooling operation computes the mean value of each 512-dimensional word embedding, yielding a 77-dimensional tensor. This tensor is passed to the 128-dimensional linear layer. Importantly, the parameters of the Vanilla text encoder are initialized randomly and updated during training.

\subsubsection{BERT Text Encoder}
The text instruction undergoes tokenization using the BERT tokenizer before feeding it into the pre-trained BERT Text Encoder \cite{bert}. The output from the BERT Text Encoder is a 768-dimensional vector, which is then passed to the 128-dimensional word embedding linear layer to obtain. Here, the weights of the BERT Text Encoder are kept frozen while only the weights to the 128-dimensional embedding layer is updated during training.

\begin{table*}[ht]
\centering
\begin{tabular}{|c|c|c|c|c|c|c|}
\hline
\textbf{Shape 1 \textbackslash{} color} & \textbf{Shape 2 \textbackslash{} color} & \textbf{Red} & \textbf{Green} & \textbf{Blue} &\textbf{Yellow} & \textbf{Black} \\
\hline
\textbf{Capsule}  & \textbf{Cube}     & \textbf{\textcolor{blue}{Test}}  & Train & Train & Train & Train \\
\hline
\textbf{Capsule}  & \textbf{Cylinder} & \textbf{\textcolor{blue}{Test}}  & Train & Train & Train & Train \\
\hline
\textbf{Capsule}  & \textbf{Prism}    & Train & Train & Train & \textbf{\textcolor{blue}{Test}}  & Train \\
\hline
\textbf{Capsule}  & \textbf{Sphere}   & Train & Train & Train & Train & \textbf{\textcolor{blue}{Test}}  \\
\hline
\textbf{Cube}     & \textbf{Cylinder} & Train & \textbf{\textcolor{blue}{Test}}  & Train & Train & Train \\
\hline
\textbf{Cube}     & \textbf{Prism}    & Train & \textbf{\textcolor{blue}{Test}}  & Train & Train & Train \\
\hline
\textbf{Cube}     & \textbf{Sphere}   & Train & Train & Train & Train & \textbf{\textcolor{blue}{Test}}  \\
\hline
\textbf{Cylinder} & \textbf{Prism}    & Train & Train & \textbf{\textcolor{blue}{Test}}  & Train & Train \\
\hline
\textbf{Cylinder} & \textbf{Sphere}   & Train & Train & \textbf{\textcolor{blue}{Test}}  & Train & Train \\
\hline
\textbf{Prism}    & \textbf{Sphere}   & Train & Train & Train & \textbf{\textcolor{blue}{Test}}  & Train \\
\hline
\end{tabular}
\caption{Train Test Split for environment C\&S\&S.}
\label{app:s3 train test split}
\end{table*}

\subsubsection{CLIP Text Encoder}
Similar to the Vanilla text encoder, the text instruction is tokenized using the CLIP tokenizer first and passed into the pre-trained CLIP Text Encoder. The resulting 512-dimensional output from the CLIP text encoder is then passed to the 128-dimensional word embedding linear layer. Similar to BERT, the parameters of the CLIP text encoder remains fixed while only the weights to the 128-dimensional embedding layer is updated during training.

\section{Text Encoder Variants Results}
\begin{figure}[ht]
  \centering
   \includegraphics[width=0.4\textwidth,height=0.3\textwidth]{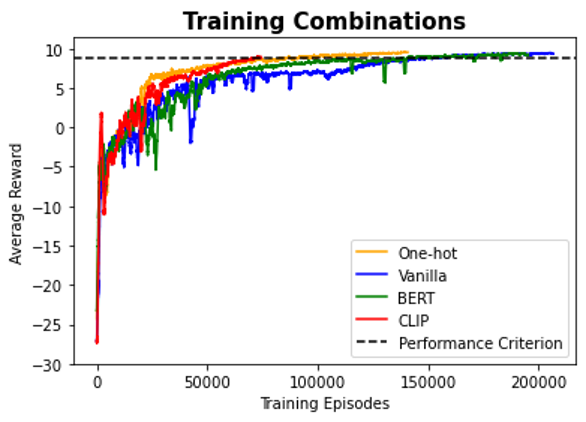}
  \hspace{0.001in}
   \includegraphics[width=0.4\textwidth,height=0.3\textwidth]{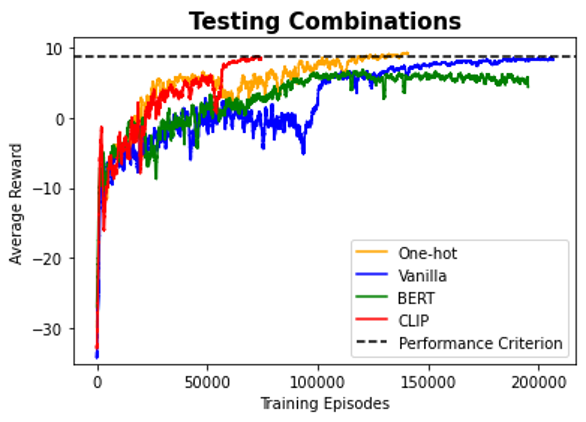}
  \caption{Learning curve of the agents with four different text encoders in training and testing combinations.}
  \label{app:training_plot}
\end{figure}

Figure \ref{app:training_plot} shows the number of episodes needed to reach performance criterion (average reward over 100 episodes $\geq$ 9.0) on the 20 training combinations and five unseen test combinations across four different text encoders: One-hot, Vanilla, BERT and CLIP. To depict the figures, we individually initialize two agents for each of the four text encoders and then compute the average of the reward during training phase from these agent pairs. The observation highlights a shared trend among all four agents, as they commence with initial average rewards of approximately -30 and steadily progress towards meeting the performance criterion.

The learning curve aligns with the results shown in the Table 1 in the main paper. The agent with CLIP text encoder reaches the performance criterion at the swiftest rate, followed by One-hot text encoder, BERT and Vanilla text encoder in training combinations. However, in testing combinations, the agent with the BERT text encoder struggles to meet the performance criterion, even after undergoing more than 200,000 training episodes. The potential reason is that BERT is solely trained on the large number of textual data, rendering it less suitable for our multi-modal learning task.

\section{PCA of Word Embeddings}
\begin{figure*}
    \centering
    \begin{subfigure}{0.33\textwidth}
        \includegraphics[width=\linewidth]{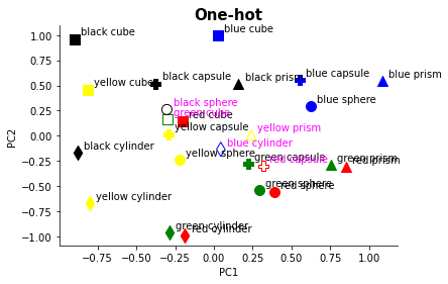}
    \end{subfigure}
    \begin{subfigure}{0.33\textwidth}
        \includegraphics[width=\linewidth]{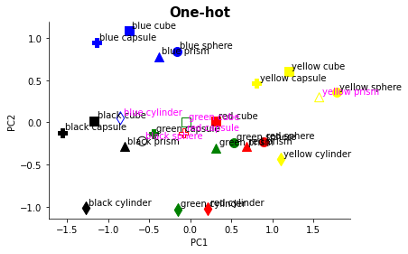}
    \end{subfigure}
    \begin{subfigure}{0.33\textwidth}
        \includegraphics[width=\linewidth]{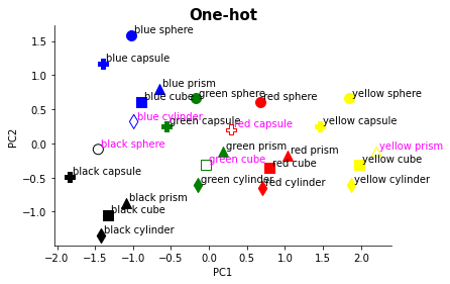}
    \end{subfigure}
    
    \begin{subfigure}{0.33\textwidth}
        \includegraphics[width=\linewidth]{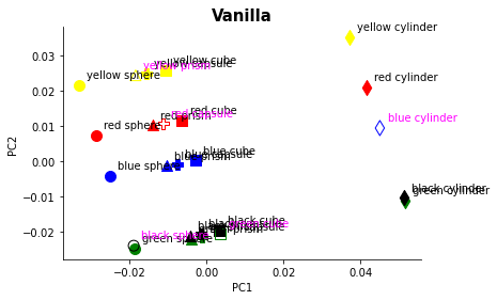}
    \end{subfigure} 
    \begin{subfigure}{0.33\textwidth}
        \includegraphics[width=\linewidth]{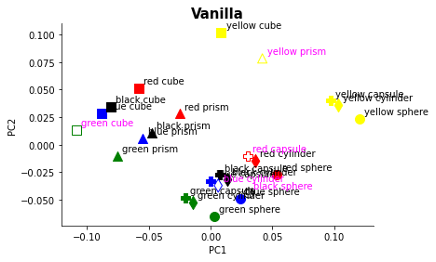}
    \end{subfigure}
    \begin{subfigure}{0.33\textwidth}
        \includegraphics[width=\linewidth]{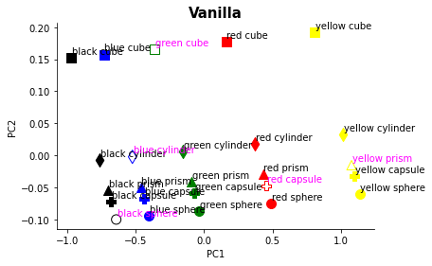}
    \end{subfigure}
    
    \begin{subfigure}{0.33\textwidth}
        \includegraphics[width=\linewidth]{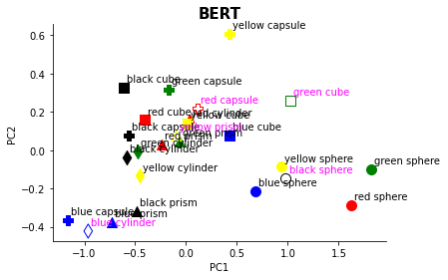}
    \end{subfigure}
    \begin{subfigure}{0.33\textwidth}
        \includegraphics[width=\linewidth]{BERT.png}
    \end{subfigure}
    \begin{subfigure}{0.33\textwidth}
        \includegraphics[width=\linewidth]{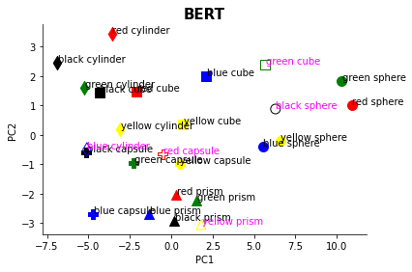}
    \end{subfigure}
    
    \begin{subfigure}{0.33\textwidth}
        \includegraphics[width=\linewidth]{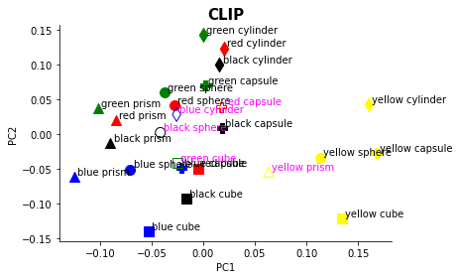}
    \end{subfigure}
    \begin{subfigure}{0.33\textwidth}
        \includegraphics[width=\linewidth]{CLIP.png}
    \end{subfigure}
    \begin{subfigure}{0.33\textwidth}
        \includegraphics[width=\linewidth]{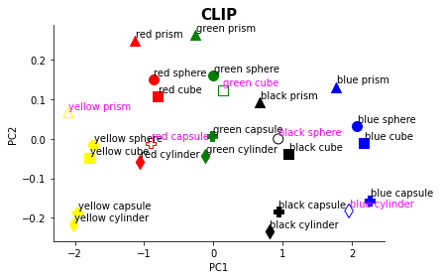}
    \end{subfigure}
    
    \caption{Word embeddings of four agents (from top to bottom: One-hot, Vanilla, BERT and CLIP) in the C\&S environment before training (left), after training 50,000 episodes (middle) and after achieving performance criterion in training combinations(right).}
    \label{app:12_word_embeddings}
\end{figure*}

In our main paper, we presented the Principal Component Analysis (PCA) results of word embeddings from the BERT and CLIP text encoders after 50,000 episodes of training. In this section, we provide a comprehensive analysis of four distinct agents within the C\&S environment. We examine their word embeddings before training, after 50,000 episodes, and upon reaching the performance criterion.

Figure \ref{app:12_word_embeddings} visually demonstrates that all the agents show the random distribution of word embeddings initially except Vanilla text encoder. The reason of Vanilla text encoder exhibits some pre-trained clusterings even before training is that it employs CLIP's tokenizer, effectively encoding the tokens for different colors and shapes. This, in turn, simplifies the process of segregating the word embeddings.

When the training episodes proceed to 50,000, the One-hot and CLIP text encoder begin to display noticeable separation among color groups and less overlaps compared to other text encoders. The figure aligns with our experimental results that the agent with the CLIP text encoder is the most efficient learner in the C\&S environment, followed by the One-hot text encoder.

Notably, BERT text encoder still did not show the clear separation between different color groups even when the agent achieves the performance criterion in training combinations. This could be the reason why the agent with BERT text encoder failed to get the average episode reward of 9 in testing combinations with the training episodes of more than 200,000. However, the agents with other three text encoders all exhibit clear separation of colors and shapes along two axes in PCA plot for both training and testing combinations.

\section{Realistic Environment Mechanics}
The learning environments were developed using the Unity 3D game engine and the Unity ML-Agents package. The interactions in the environments involve hits or collisions which are precisely managed by the Unity physics engine. This engine governs the physical interactions between the RL agent and the objects within the simulated 3D world. The integration of the Unity physics engine ensures a realistic and dynamic environment that aligns with real-world physics principles, enhancing the authenticity of the RL learning process. 


\section{Scenarios with Overlaps in Shape or Color}
\begin{figure*}[t]
    \centering
    \includegraphics[width=7in]{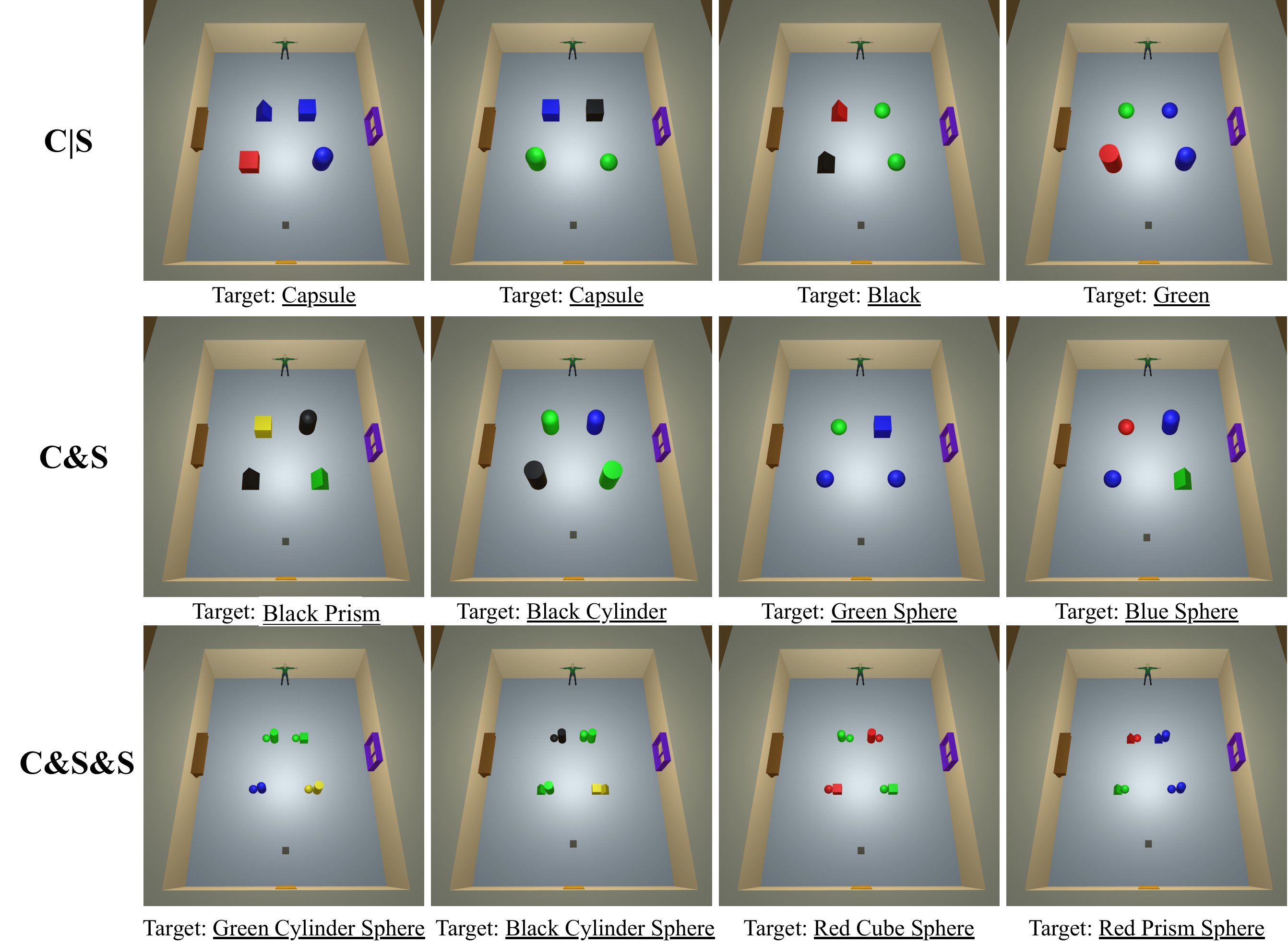} 
      \caption{These images show various scenarios of the three different environments. Of these scenarios, there are some where there could be overlaps in one characteristic, either Shape or Color.}
  \label{fig:various examples}
\end{figure*}

Figure \ref{fig:various examples} illustrates various scenarios of the three different environments. Of these scenarios, there are some where there could be overlaps in either Shape or Color.

In the first row, various instances from the C$|$S environment are presented. In this environment, the target attribute is either Shape or Color. When targeting a specific shape in C$|$S, only one object of that shape exists among the four objects. Yet, the objects may share the same color, as demonstrated by the first and second images. 

For example, in the first image, where the target instruction is "Capsule," there is only one capsule object. Since the target is shape-oriented, there is no constraint on object colors, leading to the presence of other blue-colored objects. Similarly, for color-based instructions, there is no constraint on object shapes. In the third and fourth images, there is only one object with the assigned target color, but there are other objects sharing the same shape as the target.

The second row of Figure \ref{fig:various examples} provides four examples pertaining to the C\&S environment. Here, the target instruction requires a specific combination of both shape and color. Consequently, objects sharing either the same shape or color are admissible, but there can be only one object with the precise target combination. This pattern continues in the third row, representing the C\&S\&S environment.

\begin{figure*}[ht]
    \centering
    \includegraphics[width=7in]{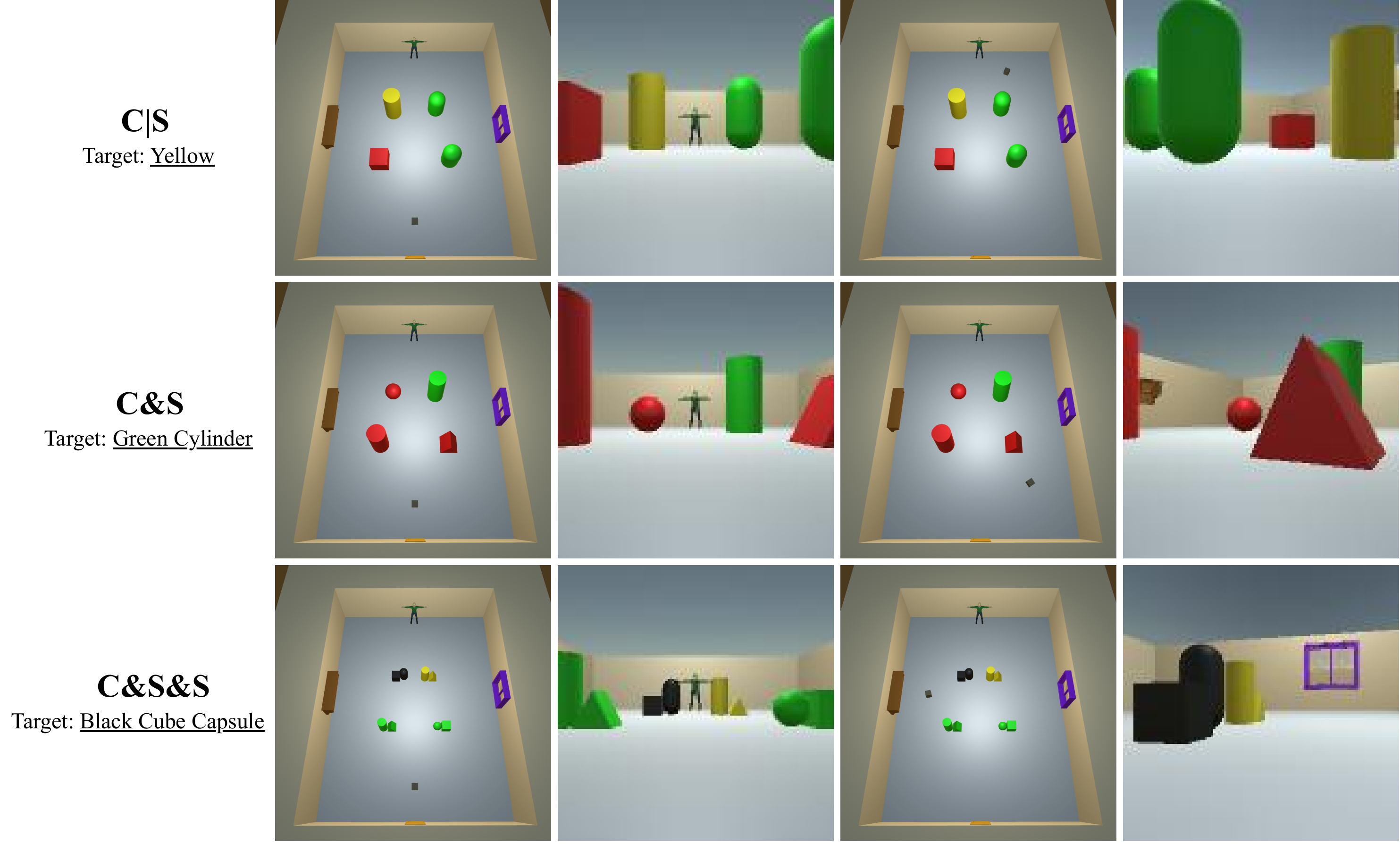} 
      \caption{Three example environments. 
      The first, second and third rows are examples from environments C$|$S , C\&S and C\&S\&S respectively. 
      The first column shows the top view of the environment, where the agent is spawned at the default starting position. 
      The second column shows the first-person view (128x128) of the RL agent in the default starting position. 
      The third column shows the top view of the environment, where the agent has moved around the environment. 
      The fourth column shows the first-person view (128x128) of the RL agent in the third column. 
      In the top view images, the agent is the small cube. For example, in the first column, the agent is the small cube located at the back of the room.      
      For the first row (C$|$S environment), the target instruction is ``yellow cylinder''.
      For the second row (C\&S environment), the target instruction is ``green cylinder''.
      For the third row (C\&S\&S environment), the target instruction is ``black cube capsule''.}
  \label{fig:all 3 envs}
\end{figure*}
\begin{figure*}
    \centering
    \begin{lstlisting}[basicstyle=\small\ttfamily]
#imports
import mlagents
from mlagents_envs.environment import UnityEnvironment as UE
from mlagents_envs.environment import ActionTuple
import numpy as np

#initialise environments
build_train_dir="S2_train_1\\build"
build_test_dir="S2_test_1\\build"
env_train =  UE(file_name=build_train_dir,seed=1,side_channels=[],worker_id=1,no_graphics = False)
env_test =  UE(file_name=build_test_dir,seed=1,side_channels=[],worker_id=2,no_graphics = False)
env_train.reset()
env_test.reset()

#sample training loop
num_episodes=5
behavior_name=list(env.behavior_specs)[0]
behavior_spec=env.behavior_specs[behavior_name]
for episode in range(num_episodes):
    decision_steps, terminal_steps = env.get_steps(behavior_name)
    tracked_agent = -1 # -1 indicates not yet tracking
    done = False # For the tracked_agent
    episode_rewards = 0 # For the tracked_agent

    while not done:
        # Note : len(decision_steps) = [number of agents that requested a decision]
        if tracked_agent == -1 and len(decision_steps) >= 1:
            tracked_agent = decision_steps.agent_id[0]
            
        # Generate an action for all agents
        action = behavior_spec.action_spec.random_action(len(decision_steps))
        # Set the actions
        env.set_actions(behavior_name, action)
        # Move the simulation forward
        env.step()
        
        # Get the new simulation results
        decision_steps, terminal_steps = env.get_steps(behavior_name)
        if tracked_agent in decision_steps: # The agent requested a decision
            episode_rewards += decision_steps[tracked_agent].reward
        if tracked_agent in terminal_steps: # The agent terminated its episode
            episode_rewards += terminal_steps[tracked_agent].reward
            done = True #set done = True since agent reach terminal state, episode ends so set to True to break out of loop
            
    print(f"Total rewards for episode {episode+1}/{num_episodes}: {episode_rewards}")
    
env.close()
    \end{lstlisting}
    \caption{Python training loop for the Unity MLAgents package.}
    \label{fig:mlagents-code}
\end{figure*}

\section{Tasks}
The primary objective of our RL agent is to navigate to the target object described by the language-based instruction. We devised three variations of the environment to investigate the agent's ability to decompose instructions and learn foundational concepts. Figure \ref{fig:all 3 envs} shows an example from these three environments.
 
The first environment is named ``{Color} {Shape}''(C\&S) as the agent is only rewarded when it navigates to the target that satisfies both color and shape descriptions, and is penalised if it navigates to any of the other three objects that either have different color or shape attributes. For example, when the instruction given to the agent is ``red cube'', the agent has to learn to navigate to the corresponding object that has the red color and cube shape combination. Navigating to either ``red sphere'' or ``blue cube'' will disburse a negative reward. 

The second environment is named ``{Color} {Shape} {Shape}'' (C\&S\&S) as the agent's goal is to navigate to one of the 4 predetermined locations that contain two objects described by a color attribute and two distinct shape combinations. Examples of the color-shape-shape combination instructions are ``red cube cylinder'' or ``green sphere prism''. This requires the agent to compose not just color and shape concepts but to compose three attributes to be rewarded.

The third environment is named ``Color'' OR ``Shape''(C$|$S) as the agent's goal is to learn shape or color invariances respectively. For example, if the rewarded target is ``red cube'', the instruction given to the agent will either be ``red'' or ``cube'' and not the combination of both attributes. The agent can navigate to any red object or any colored cube to be rewarded respectively. This motivates the agent to learn shape or color invariances as two distinct concepts, which is different from the compositional learning goal of the first two environments. 



To summarise, these are the key details of the three environments:
 \begin{enumerate}
     \item  C\&S (``Color AND Shape''): The task is to compositionally learn two word concepts. There is one target described by the color and shape attribute with three non-target objects. The instruction includes both color and shape words as inputs.
     \item  C\&S\&S (``Color Shape Shape''): The task is to compositionally learn three-word concepts. There is one target with two shapes and three non-targets with pairs of shapes. Both objects in a pair will be of the same color. The instruction includes all three color and two shape words as inputs. The train test split of this environment is shown in Table \ref{app:s3 train test split}.    
      \item  C$|$S (``Color'' OR ``Shape''): The task is single word concept learning. There is one target and three non-target objects. Only one words, either color or shape, that is given as instruction.
 \end{enumerate}
\end{document}